\documentclass[twoside,11pt]{article}

\usepackage{enumerate}
\usepackage{jmlr2e}
\usepackage{amsmath}
\usepackage{amssymb}
\usepackage{graphicx}
\usepackage{accents}
\usepackage{pslatex}
\usepackage{url}
\usepackage{bm}
\usepackage{color}
\usepackage{ulem}
\usepackage{cancel}

\definecolor{burntorange}{rgb}{0.8, 0.33, 0.0}
\definecolor{cool_green}{rgb}{0.0, 0.5, 0.0}

\jmlrheading{0}{0}{0}{0}{0}{???}

\ShortHeadings{Vicinal Risk Minimization}{}
\firstpageno{0}

\begin{document}

\title{Generalization Bounds for Vicinal Risk Minimization Principle}

\author{\name Chao Zhang
\email chao.zhang@dlut.edu.cn \\
       \addr  School of Mathematical Sciences\\
       Dalian University of Technology\\
       Dalian, Liaoning 116024, P.R China
       \AND
       \name Min-Hsiu Hsieh\footnote{Corresponding author}
       \email min-hsiu.hsieh@uts.edu.au\\
       \addr Centre for Quantum Software and Information\\
University of Technology, Sydney\\
 Ultimo NSW 2007, Australia
         \AND
       \name Dacheng Tao\email dacheng.tao@sydney.edu.au \\
       \addr School of Computer Science\\
       The University of Sydney\\
       Darlington, NSW 2008, Australia}

\editor{}

\maketitle

\begin{abstract}
The vicinal risk minimization (VRM) principle, first proposed by \citet{vapnik1999nature}, is an empirical risk minimization (ERM) variant that replaces Dirac masses with vicinal functions. Although there is strong numerical evidence showing that VRM outperforms ERM if appropriate vicinal functions are chosen, a comprehensive theoretical understanding of VRM is still lacking. In this paper, we study the generalization bounds for VRM. Our results support Vapnik's original arguments and additionally provide deeper insights into VRM. First, we prove that the complexity of function classes convolving with vicinal functions can be controlled by that of the original function classes under the assumption that the function class is composed of Lipschitz-continuous functions. Then, the resulting generalization bounds for VRM suggest that the generalization performance of VRM is also effected by the choice of vicinity function and the quality of function classes. These findings can be used to examine whether the choice of vicinal function is appropriate for the VRM-based learning setting. Finally, we provide a theoretical explanation for existing VRM models, {\it e.g.}, uniform distribution-based models, Gaussian distribution-based models, and mixup models.

\end{abstract}

\begin{keywords}
vicinal risk minimization, generalization bound, statistical learning theory, empirical risk minimization
\end{keywords}

\section{Introduction}\label{sec:intr}

Let $\mathcal{X}\subset \mathbb{R}^I$ and $\mathcal{Y}\subset \mathbb{R}^J$ be an input space and the corresponding output space, respectively. Given a function class $\mathcal{G}\subset \mathcal{Y}^{\mathcal{X}}$ and a loss function $\ell$, it is expected to find a function $g_*\in\mathcal{G}$ to minimize the expected risk over $\mathcal{G}$:
\begin{align}\label{eq:exrisk}
\mathrm{R} (\ell\circ g):=&\int \ell(g({\bf x}),{\bf y})d\mathrm{P}({\bf x},{\bf y}),
\end{align}
where $\ell\circ g$ is the composition of $\ell$ and $g$, and $\mathrm{P}({\bf x},{\bf y})$ stands for the joint distribution of $({\bf x},{\bf y})\in\mathcal{X}\times\mathcal{Y}$.

Since $\mathrm{P}({\bf x},{\bf y})$ usually is unknown, it is difficult to obtain the solution $g_*$ by directly minimizing the expected risk $\mathrm{R}(\ell\circ g)$. Instead, the empirical risk minimization (ERM) principle is adopted to handle this issue: given a sample set $\{({\bf x}_n,{\bf y}_n)\}_{n=1}^N$, one minimizes the empirical risk over $\mathcal{G}$
\begin{equation}\label{eq:emrisk}
\widehat{\mathrm{R}}(\ell\circ g):=\frac{1}{N}\sum_{n=1}^N \ell(g({\bf x}_n),{\bf y}_n),
\end{equation}
to find an approximation of $g_*$:
\begin{equation}\label{eq:emsolution}
\widehat{g}:=\arg\min_{g\in\mathcal{G}}\widehat{\mathrm{R}}(\ell\circ g).
\end{equation}
For ERM learning processes, one of major research concerns is their generalization performance, {\it i.e.,} whether the learning result $\widehat{g}$ performs well on the new inputs that are not included in the training set. There have been intensive works on the generalization bounds for ERM \citep[see][]{bousquet2002stability,bartlett2005local,zou2014generalization,mohri2010stability} and the list provided here is incomplete.


\subsection{Vicinal Risk Minimization Principle}

\citet{vapnik1999nature} has pointed out that the empirical risk \eqref{eq:emrisk} can be equivalently rewritten as
\begin{equation}\label{eq:emrisk.delta}
\widehat{\mathrm{R}}(\ell\circ g)=\int \ell(g({\bf x}),{\bf y}) d\widehat{\mathrm{P}}({\bf x},{\bf y})
\end{equation}
with the empirical distribution
\begin{equation}\label{eq:emfunal}
\widehat{\mathrm{P}}({\bf x},{\bf y}) = \frac{1}{N}\sum_{n=1}^N\delta_{{\bf x}_n}({\bf x})\cdot\delta_{{\bf y}_n}({\bf y}),
\end{equation}
where $\delta_{{\bf a}}({\bf x})$ stands for the Dirac mass centered at the point ${\bf a}$. Since the support of $\delta_{{\bf x}_n}({\bf x})$ is an one-point set $\{{\bf x}_n\}$ ($n=1,2,\cdots,N$), $\widehat{\mathrm{P}}({\bf x},{\bf y})$ cannot approximate $\mathrm{P}({\bf x},{\bf y})$ accurately if $\mathrm{P}({\bf x},{\bf y})$ is continuous. Thus, the vicinal risk minimization (VRM) principle was proposed to improve the feasibility of ERM by replacing the Dirac mass with some vicinity function defined on the set $\Omega({\bf x}_n) := \{ {\bf x}: d({\bf x} , {\bf x}_n) \leq r_{{\bf x}_n}   \}$ w.r.t. the metric $d(\cdot,\cdot)$, where the set $\Omega({\bf x}_n)$ is called the {\bf vicinity} of the point ${\bf x}_n$.

As suggested by \citet{vapnik1999nature}, the Dirac mass function $\delta_{{\bf x}_n}({\bf x})$ in \eqref{eq:emfunal} can respectively be replaced with two types of vicinity functions: the uniform distribution function with the density $p({\bf x} | \Omega({\bf x}_n)) = [{\rm vol}(\Omega({\bf x}_n))]^{-1}$ and the Gaussian distribution function $\mathcal{N}({\bf x}_n,\bm{\Sigma}({\bf x}_n))$. Recently, \citet{zhang2017mixup} replaced the product $\delta_{{\bf x}_n}({\bf x})\cdot\delta_{{\bf y}_n}({\bf y})$ in \eqref{eq:emfunal} with
the mixup vicinity function:
\begin{align}\label{eq:mixup}
{\rm V}_{\rm mix}({\bf x},{\bf y}\,|\,{\bf x}_n,{\bf y}_n):=
\frac{1}{N}\sum_{m=1}^N \mathbb{E}_\lambda \big\{ \delta_{\lambda{\bf x}_n+(1-\lambda){\bf x}_m}({\bf x}) \cdot\delta_{\lambda{\bf y}_n+(1-\lambda){\bf y}_m}({\bf y}) \big\},
\end{align}
where $\lambda\sim {\rm Beta}(\alpha,\alpha)$ and $\alpha>0$. {Their numerical experiments showed that the introduction of mixup vicinity function improves the performance of the state-of-art deep models ({\it e.g.,} ResNet, ResNeXt and VGG) on several well-known datasets including ImageNet-2012, CIFAR-10, CIFAR-100, Google commands and UCI datasets.}

In this work, we consider the VRM principle for which {the vicinal risk has the following general form:} 
\begin{align}\label{eq:vrisk}
\mathrm{R}_\nu (\ell\circ g) := \int \ell(g({\bf x}),{\bf y}) d \mathrm{P}_\nu ({\bf x},{\bf y})
\end{align}
with the vicinal distribution
 \begin{equation}
\mathrm{P}_\nu ({\bf x},{\bf y}) := \frac{1}{N}\sum_{n=1}^N {\rm V}\big({\bf x},{\bf y} | \theta({\bf x}_n,{\bf y}_n)\big ) ,
\end{equation}
where ${\rm V}\big({\bf x},{\bf y} | \theta({\bf x}_n,{\bf y}_n)\big )$ is the vicinity function at the point $({\bf x},{\bf y})$ with the parameter $\theta({\bf x}_n,{\bf y}_n)$. The form \eqref{eq:vrisk} covers the aforementioned three types of VRM models as special cases.

\citet{chapelle2001vicinal} addressed that the VRM principle can be understood from two equivalent viewpoints:
 \begin{itemize}
\item[(V1)] The vicinal distribution $\mathrm{P}_\nu ({\bf x},{\bf y})$, which is the convolution of the empirical distribution $\widehat{\mathrm{P}}({\bf x},{\bf y})$ with the vicinal function ${\rm V}$, can be deemed as an approximation to the underlying distribution {$\mathrm{P}({\bf x},{\bf y})$}. If $\mathrm{P}_\nu ({\bf x},{\bf y})$ approximates $\mathrm{P}({\bf x},{\bf y})$ accurately, VRM can still perform well in spite of the fact that $\mathcal{F}$ is not chosen with suitably small capacity.

\item[(V2)] For any ${\bf a} \in\mathcal{X}\times \mathcal{Y}$ and any $g\in\mathcal{G}$, denote
\begin{equation}\label{eq:phi}
\phi(\ell\circ g,{\bf a}) := \int  \ell(g({\bf x}),{\bf y}) d{\rm V}({\bf x},{\bf y}|\theta({\bf a})).
\end{equation}
Then, the vicinal risk $\mathrm{R}_\nu (\ell\circ g)$ can be equivalently rewritten as
\begin{equation}\label{eq:emvrisk}
\mathrm{R}_\nu (\ell\circ g) = \frac{1}{N}\sum_{n=1}^N \phi\big(\ell\circ g,({\bf x}_n,{\bf y}_n)\big) = \widehat{\mathrm{R}}\big(\phi\big(\ell\circ g,({\bf x}_n,{\bf y}_n)\big)\big),
\end{equation}
which means that VRM actually modifies the error value $\ell(g({\bf x}_n),{\bf y}_n)$ by convolving it with the vicinity function ${\rm V}$. In this way, the VRM principle for the function class $\mathcal{G}$ turns out to be the ERM principle for the function class $\{\phi(\ell\circ g,({\bf x},{\bf y})):g\in\mathcal{G}\}$.
\end{itemize}

Although many experimental evidences support the superiority of VRM \citep[see][]{chapelle2001vicinal,cao2015use,ni2015training,hai2010vicinal,zhang2017mixup}z, to the best of our knowledge, VRM research beyond that is still very limited.

\subsection{Generalization Bounds for VRM}\label{sec:motivation}

For convenience, denote $\mathcal{Z} =\mathcal{X}\times \mathcal{Y}\subset \mathbb{R}^{K} $ with $K=I+J$ and ${\bf z}= ({\bf x},{\bf y})\in\mathbb{R}^{K}$. Let $D(\mathcal{Z})$ be the set of {distributions} on $\mathcal{Z}$. Denote $f({\bf z})=\ell(g({\bf x}),{\bf y})$ and define $\mathcal{F}:=\{{\bf z} \mapsto \ell(g({\bf x}),{\bf y}):g\in \mathcal{G}\}$. 
To ease the notation in the following discussion, we denote 
\begin{align*}
&\mathrm{R} f = \int f({\bf z})d\mathrm{P}({\bf z}) := \mathrm{R} (\ell\circ g),   \\
&\widehat{\mathrm{R}} f =\frac{1}{N}\sum_{n=1}^N f({\bf z}_n) := \widehat{\mathrm{R}} (\ell\circ g),  \\
&\mathrm{R}_\nu f =\int f({\bf z}) d \mathrm{P}_\nu ({\bf z}) := \mathrm{R}_\nu (\ell\circ g),   
\end{align*}
and let $f_\nu $ be the solution to the minimization of the vicinal risk $\mathrm{R}_\nu f$ over $\mathcal{F}$. In a VRM learning process, the quantity $\mathrm{R}f_\nu -\mathrm{R}_\nu f_\nu $
corresponds to the estimation of the expected risk of $f_\nu $ from its vicinal risk. Before seeing the samples, we do not know what function will be selected by the VRM-based algorithm. Therefore, we need to consider the uniform deviation
\begin{equation}\label{eq:def.bound}
\mathrm{R}f_\nu -\mathrm{R}_\nu f_\nu \leq \sup_{f\in\mathcal{F}}\big\{\mathrm{R}f-\mathrm{R}_\nu f\big\},
\end{equation}
which is the so-called generalization bound for VRM. As shown in \eqref{eq:def.bound}, VRM's generalization performance is influenced by at least two factors: the choice of vicinity function ${\rm V}$ and the complexity of function class $\mathcal{F}$. In the following discussion, we will show that VRM's generalization performance is also influenced by the quality of the function class, which differs from the scenario of ERM (see Section \ref{sec:omega}).

To  apply known results from statistical learning theory, one can proceed to relax the right-hand side of \eqref{eq:def.bound} as follows. It follows from \eqref{eq:phi} and \eqref{eq:emvrisk} that
\begin{align}\label{eq:rhs1}
\sup_{f\in\mathcal{F}}\big\{\mathrm{R}f-\mathrm{R}_\nu f\big\}\leq \sup_{f\in\mathcal{F}}\big\{\mathrm{R}f-\mathrm{R}
\phi(f,{\bf z})\big\}+\sup_{f\in\mathcal{F}}\big\{\mathrm{R}\phi(f,{\bf z})-\widehat{\mathrm{R}} \phi(f,{\bf z}_n)\big\},
\end{align}
where $\phi(f,{\bf z}):=   \phi\big(\ell\circ g,({\bf x},{\bf y})\big)$, $\mathrm{R}
\phi(f,{\bf z}) := \int \phi(f,{\bf z}) d \mathrm{P} ({\bf z})$ and $\widehat{\mathrm{R}}
\phi(f,{\bf z}_n) := \frac{1}{N}\sum_{i=1}^N \phi(f,{\bf z}_i) $. The first term in the right-hand side of \eqref{eq:rhs1} can be used to analyze the effect caused by the introduction of VRM. The second term actually is the generalization bound of the ERM learning process for the function class $\{\phi(f,{\bf z}):f\in\mathcal{F}\}$, and thus can be bounded by using the classical results on generalization bounds for ERM based on various complexities of function classes, {\it e.g.,} Vapnik-Chervonenkis (VC) dimensions \citep{vapnik1999nature}, covering numbers \citep{mendelson2003few} and Rademacher complexities \citep{bousquet2004introduction}. However, since the supreme operation in the term $\sup_{f\in\mathcal{F}}\big\{\mathrm{R}f-\mathrm{R}
\phi(f,{\bf z})\big\}$ is taken over the entire function class $\mathcal{F}$, it will be difficult to analyze the behavior of the term when the complexity of $\mathcal{F}$ is high. To overcome this limitation, we will use the combinatorial method, presented by \citet{pollard1984convergence}, to achieve the upper bound of $\sup_{f\in\mathcal{F}}\{\mathrm{R}f-\mathrm{R}_\nu f\}$, and then explore VRM's inherent characteristics based on a countable subset of $\mathcal{F}$ (see Proposition \ref{pro:analysis}).

\subsection{Overview of Main Results}

In this paper, we study the generalization bounds for VRM and then analyze factors influencing VRM's generalization performance. In particular, we prove that the complexity of the specific function class for VRM can be controlled by that of the original function classes. Therefore, VRM's generalization performance is mainly determined by the choice of vicinal functions and the quality of function classes. Our theoretical findings support Vapnik's original arguments about VRM presented by \citet{vapnik1999nature}: 1) if the vicinal functions are not suitably chosen, VRM still performs well if the function class $\mathcal{F}$ has low complexity; and 2) even if complexity of $\mathcal{F}$ is not suitably small, VRM still has satisfactory performance if the vicinal functions are suitably chosen. Moreover, we introduce the quantity $\Omega_\nu$, defined in \eqref{eq:U}, to measure the effectiveness of introducing vicinal functions and obtain a strategy to examine whether the choice of vicinal functions is suitable for the VRM-based learning setting. Finally, we provide a theoretical explanation of some existing VRM models including uniform distribution-based models, Gaussian distribution-based models, and mixup models.

In contrast to classical statistical learning theory, obtaining the generalization bounds for VRM requires additional considerations. In view of VRM's inherent characteristics, we introduce the concept of {\it vicinity ghost samples} to exploit the positional relationship between samples and their ghosts. Then, we analyze the behavior of the sample pair composed of one sample and the ghost sample lying in its vicinity. In addition, to guarantee the existence of such a sample pair, we prove that the probability that the Euclidean distance between the sample and its vicinity ghost is large will exponentially decay to {\it zero} when the sample size $N$ goes to {\it infinity}.

Based on the sample pairs, we introduce the {\it difference function class} {$\mathcal{P}:=\{f({\bf z}_1)-\phi(f,{\bf z}_2):f\in\mathcal{F},\;{\bf z_1},{\bf z_2}\in\mathcal{Z}\}$}, and then
study the complexity of $\mathcal{P}$. We prove that the covering number of $\mathcal{P}$ can be controlled by that of $\mathcal{F}$. To obtain a distribution-free upper bound of the covering number of $\mathcal{P}$, we then develop two types of uniform entropy numbers (UENs) for $\mathcal{P}$: the first by selecting $N$ points from $\mathcal{Z}\times\mathcal{Z}$ and the second by selecting elements from a $\xi$-rectangle cover of $\mathcal{Z}\times\mathcal{Z}$. We highlight that the complexity of $\mathcal{P}$ is dominated by the latter UEN, which actually is defined based on the positional relationship between samples and their ghosts. This result also implies that the positional relationship plays an important role in studying VRM's properties.

Since $\mathbb{E}_{{\bf Z} \sim (D(\mathcal{Z}))^N}\{\widehat{\mathrm{R}}f\}$ usually is not identical to $\mathbb{E}_{{\bf Z} \sim (D(\mathcal{Z}))^N}\{\mathrm{R}_vf\}$, we will apply the one-sided concentration inequality for the random variables with non-zero means. Moreover, we present the specific symmetrization inequality for VRM. Although the symmetrization inequality is of the similar form with that of the classical symmetrization for ERM, it has some specific considerations.

The rest of this paper is organized as follows. In Section \ref{sec:ghost}, we introduce the concept of vicinity ghost samples and Section \ref{sec:uen} studies the complexity of difference function classes. In Section \ref{sec:main}, we present the generalization bounds for VRM and the last section concludes the paper. In Appendix~\ref{sup:sym}, we present the symmetrization and the concentration inequalities respectively. The proofs of main results are given in Appendix~\ref{sup:proof}.




\section{Vicinity Ghost Samples}\label{sec:ghost}

Let $\Pi$ be the collection of all permutations of the set $\{1,\dots,N\}$. Given an i.i.d.~sample set ${\bf Z}= \{{\bf z}_1,\cdots,{\bf z}_N\}$, let ${\bf Z}'=\{{\bf z}'_1,\cdots,{\bf z}'_N\}$ be its ghost sample set. It is noteworthy that
\begin{align}\label{eq:permute}
\widehat{\mathrm{R}}'f-\mathrm{R}_\nu f = \frac{1}{N}\sum_{n=1}^N\big( f({\bf z}'_n) - \phi(f,{\bf z}_n)  \big)
=  \frac{1}{N}\sum_{n=1}^N\big( f({\bf z}'_{\pi(n)}) - \phi(f,{\bf z}_n)  \big),
\end{align}
where $\pi\in\Pi$ is any permutation.
In view of VRM's characteristic, we would like to study the behavior of $f({\bf z}'_{\pi(n)})-\phi(f,{\bf z}_n)$ when ${\bf z}'_{\pi(n)}$ locates in the vicinity of ${\bf z}_n$. For this purpose, we introduce the concept of {\it vicinity ghost samples} to exploit the positional relationship between samples and their ghosts. We then present a theoretical guarantee that for any sample ${\bf z}_{n}\in {\bf Z}$, the corresponding ghost sample will locate in its vicinity with a high probability.

\begin{definition}[Vicinity Ghost Samples]\label{def:vgs}
Given a sample set ${\bf Z}= \{{\bf z}_1,\cdots,{\bf z}_N\}$ and its ghost set ${\bf Z}'=\{{\bf z}'_1,\cdots,{\bf z}'_N\}$, let
%
\begin{equation}\label{eq:permute}
\pi_* = \mathop{\arg\min}_{\pi\in\Pi} \sum_{n=1}^N \|{\bf z}'_{\pi(n)}-{\bf z}_n \|_2.
\end{equation}
The resulted sequence {${\bf Z}_{\pi_*}^\dagger:=\{{\bf z}'_{\pi_*(1)},\cdots,{\bf z}'_{\pi_*(N)}\}$}  is called the vicinity ghost samples. Moreover, denote the sample pair {${\bf s}_{[n]} := ({\bf z}'_{\pi_*(n)},{\bf z}_n)$} and ${\bf S}^\dag:= \{ {\bf s}_{[1]},\cdots,{\bf s}_{[N]}\}$.

\end{definition}
There are two things worth noting: 1) the sample pairs ${\bf s}_{[1]},\cdots,{\bf s}_{[N]}$ are not independent any more; and 2) {the point ${\bf z}'_{\pi_*(n)}$ could be far from  ${\bf z}_n$ in some cases, for example, when} they locate in the tail of a distribution on $\mathcal{Z}$. Therefore, it is critical to show that the point ${\bf z}'_{\pi_*(n)}$ locates in the vicinity of the counterpart ${\bf z}_n$ with high probability. {One main challenge is to provide a bridge between the Euclidean distance and the probability distribution. Here, we introduce a probability-based distance to measure the difference between two points in $\mathbb{R}^K$ and then prove that the probability that ${\bf z}'_{\pi_*(n)}$ is far from ${\bf z}_n$ w.r.t. this distance will exponentially decay to {\it zero} as the sample size $N$ approaches {\it infinity}. 

\begin{definition}[CDF Distance]\label{def:CDFdistance}
Let $\bf{F}({\bf t})$ be the cumulant density function (cdf) of a distribution on $\mathcal{Z}\subset\mathbb{R}^K$:
\begin{equation*}
{\bf F}({\bf t}): =\int_{(-\infty,{\bf t}]} d \mathrm{P}({\bf z}),\quad  \forall \,{\bf t}=(t^{(1)},\cdots,t^{(K)})\in\mathbb{R}^K,
\end{equation*}
where $(-\infty,{\bf t}]$ denotes the set $(-\infty,t^{(1)}]\times\cdots\times (-\infty,t^{(K)}]$, and denote its $k$-th marginal cdf as 
\begin{equation*}
{\rm F}_k({\bf t}): = \lim_{t^{(1)},\cdots,t^{(k-1)},t^{(k+1)},\cdots,t^{(K)}\rightarrow +\infty} {\rm F}({\bf t}).
\end{equation*}
The cdf distance between two points ${\bf t}_1, {\bf t}_2\in\mathbb{R}^K$ is defined as
\begin{equation*}
d_{\mathrm{F}}({\bf t}_1, {\bf t}_2) : = \sqrt{\sum_{k=1}^K \big(\mathrm{F}_k ({\bf t}_1) -\mathrm{F}_k ({\bf t}_2)\big)^2 }.
\end{equation*}
\end{definition}
It is easy to prove that $d_{\mathrm{F}}({\bf t}_1, {\bf t}_2)$ satisfies non-negativity, symmetry and the triangle inequality. The details on cdf distance are referred to Chapter 4.2 of \citet{venturini2015statistical}. Since the distribution $\mathrm{P}({\bf z})$ usually is unknown, there is also an empirical version of the cdf distance:
\begin{definition}[Empirical CDF Distance]
Let ${\bf Z}= \{{\bf z}_1,\cdots,{\bf z}_N\}$ be an i.i.d.~sample set taken from a distribution on $\mathcal{Z}$ with ${\bf z}_n = (z_n^{(1)},\cdots,z_n^{(K)})$. Denote $\widehat{\mathrm{F}}({\bf t})$ as the empirical cdf: 
\begin{equation*}
\widehat{\mathrm{F}}({\bf t}):=
\frac{1}{N} \sum_{n=1}^N  {\bf 1}_{(-\infty,{\bf t}]}({\bf z}_n),\;\; \forall \,{\bf t}=(t^{(1)},\cdots,t^{(K)})\in\mathbb{R}^K,
\end{equation*}
and denote $\widehat{\mathrm{F}}_k ({\bf t})$ as the $k$-th marginal empirical cdf:
\begin{equation*}
\widehat{\mathrm{F}}_k ({\bf t}) :=  \frac{1}{N} \sum_{n=1}^N  {\bf 1}_{(-\infty,t^{(k)}]}(z^{(k)}_n).
\end{equation*}
The empirical cdf distance between two points ${\bf t}_1, {\bf t}_2\in\mathbb{R}^K$ is defined as
\begin{equation}\label{eq:ecdf}
d_{\widehat{\mathrm{F}}}({\bf t}_1, {\bf t}_2) : = \sqrt{\sum_{k=1}^K \big(\widehat{\mathrm{F}}_k ({\bf t}_1) -\widehat{\mathrm{F}}_k ({\bf t}_2)\big)^2 }.
\end{equation}
 
\end{definition}

As mentioned in \citet{venturini2015statistical}, it follows from the strong law of large numbers that the empirical cdf distance $d_{\widehat{\mathrm{F}}}({\bf t}_1, {\bf t}_2)$ will converge to the cdf distance $d_{\mathrm{F}}({\bf t}_1, {\bf t}_2)$.

\begin{theorem}\label{thm:point}

Let ${\bf Z}$ and ${\bf Z}^\dag$ be a sample set of $N$ elements and its vicinity ghost set taken from a distribution on $\mathcal{Z}\subset \mathbb{R}^K$, respectively. Then, for any $\xi>0$,
\begin{align*}
&\mathbb{P}\left\{ \max_{n\in\{1,\cdots,N\}}\big[d_{\widehat{\mathrm{F}}}({\bf z}_n, {\bf z}_{\pi_*(n)})\big]>\xi\right\}
\leq  c \cdot {\rm e}^{\frac{-N\xi^2}{2 K}},
\end{align*}
where $c>0$ is an absolute constant.
\end{theorem}
This theorem shows that the probability that the empirical cdf distance between the sample ${\bf z}_n$ and its vicinity ghost ${\bf z}_{\pi_*(n)}$ is larger than a positive constant $\xi$ will exponentially decay to ${\it zero}$ when the sample size $N$ approaches {\it infinity}. Let $\mathcal{C}_{r}({\cal Z})$ be a cover of ${\cal Z}$ at radius $r>0$ w.r.t. the empirical distance $d_{\widehat{\mathrm{F}}} (\cdot,\cdot)$. Define the mutually exclusive events:
\begin{itemize}
\item $\mathcal{E}_1$ = ``For any sample pair $({\bf z}'_{\pi_*(n)},{\bf z}_{n})$, there exists an element $C\in\mathcal{C}_{r}({\cal Z})$ that contains ${\bf z}'_{\pi_*(n)}$ and ${\bf z}_n$;"
 \item $\mathcal{E}_2$ = ``There always exists at least one pair of samples ${\bf z}'_{\pi_*(n)}$ and ${\bf z}_{n}$ that cannot be simultaneously contained by any individual element $C\in\mathcal{C}_{r}({\cal Z})$."
\end{itemize}
It follows from  
\begin{equation*}
\mathbb{P} \{ \mathcal{E}_2\}=\mathbb{P}\left\{ \max_{n\in\{1,\cdots,N\}}\big[ d_{\widehat{\mathrm{F}}}({\bf z}_n, {\bf z}_{\pi_*(n)})\big]>r\right\}
\end{equation*}
that
\begin{equation*}
\mathbb{P}\{\mathcal{E}_1 \} = 1-\mathbb{P}\{\mathcal{E}_2 \}\geq1- c \cdot {\rm e}^{\frac{-Nr^2}{2 K}},
\end{equation*}
which implies that the sample pair $({\bf z}'_{\pi_*(n)},{\bf z}_{n})$ can be contained by an element $C\in\mathcal{C}_r(\mathcal{Z})$ with overwhelming probability, i.e., that the event $\mathcal{E}_1$ does not hold decays exponentially to zero when the sample size $N$ approaches {\it infinity}.}



\section{Difference Function Classes}\label{sec:uen}

For any $f\in\mathcal{F}$, denote the difference function $p({\bf s}):=f({\bf z}_1)-\phi(f,{\bf z}_2)$ with ${\bf s}=({\bf z}_1,{\bf z}_2)\in\mathcal{Z}\times\mathcal{Z}$. 
Moreover, define the difference function class:
\begin{equation}\label{eq:P}
\mathcal{P}:=\{f({\bf z}_1)-\phi(f,{\bf z}_2):f\in\mathcal{F}\}.
\end{equation}

Given a sample set ${\bf Z}=\{{\bf z}_{1},\cdots,{\bf z}_{N}\}$ and its vicinity ghost ${\bf Z}^\dag=\{{\bf z}'_{\pi_*(1)},\cdots,{\bf z}'_{\pi_*(N)}\}$, we then have
\begin{equation*}
\mathbb{P}\Big\{\sup_{f\in\mathcal{F}}\widehat{\mathrm{R}}'f-\mathrm{R}_\nu f>\frac{\xi}{2}\Big\}= \mathbb{P}\Big\{\sup_{p\in\mathcal{P}}\frac{1}{N}\sum_{n=1}^N p({\bf s}_{[n]})>\frac{\xi}{2}\Big\}.
\end{equation*}
To obtain an upper bound of the above probability, the supremum $\sup_{p\in\mathcal{P}}$ should be relaxed to be the summation operation by using the complexity measure of $\mathcal{P}$. It is also one of main research issues of statistical learning theory. In this paper, we are mainly concerned with the covering number of the difference function class $\mathcal{P}$, and refer to \citet{mendelson2003few,zhou2003capacity} for further details.
\begin{definition}\label{def:covnum}
{Let $({\cal M}, d)$ be a metric space and $\mathcal{F}$ be a subset of ${\cal M}$. A subset $\Lambda$ of ${\cal F}$ is said to be a cover of ${\cal F}$ if for each $f\in \mathcal{F}$ there exists $h\in \Lambda$ such that $d(f,h)\leq \xi$. The covering number, denote $\mathcal{N}(\mathcal{F},\xi,d)$, is the minimum cardinality of any cover of $\mathcal{F}$. }
\end{definition}

Note that the function $\phi(f,{\bf z})$ is derived by convolving $f({\bf z})$ with the vicinity function. Intuitively, the complexity of $\mathcal{P}$ should not be higher than that of $\mathcal{F}$. The following result reveals the relationship between the covering numbers $\mathcal{N}(\mathcal{F},\xi,\ell_1({\bf Z}))$ and $\mathcal{N}(\mathcal{P},\xi,\ell_1({\bf S}^\dag))$, where $\ell_1({\bf Z})$ stands for the $\ell_1$-norm defined on the set ${\bf Z}$.
\begin{theorem}\label{thm:covering}
Given a function class $\mathcal{F}$ and a function $\phi:\mathcal{F}\times \mathcal{Z} \rightarrow \mathbb{R}$, if $\phi$ satisfies the $\lambda$-Lipschitz condition that $|\phi(f,{\bf z}) - \phi(h,{\bf z})|\leq \lambda |f({\bf z})-h({\bf z})|$, $\forall f,h\in\mathcal{F}$, then it holds that for any $\xi>0$,
\begin{equation*}
\mathcal{N}\big(\mathcal{P},\xi,\ell_1({\bf S}^\dag) \big)\leq \mathcal{N}\Big(\mathcal{F},\frac{\xi}{2+\lambda},\ell_1({\bf Z}) \Big),
\end{equation*}
where ${\cal P}$ and ${\bf S}^\dag$ are defined in \eqref{eq:P} and Definition \ref{def:vgs}, respectively. Furthermore, if $\lambda\in(0,1)$, then it holds that
\begin{equation*}
\mathcal{N}\Big(\mathcal{F},\frac{\xi}{1-\lambda},\ell_1({\bf Z}) \Big)\leq \mathcal{N}\big(\mathcal{P},\xi,\ell_1({\bf S}^\dag) \big).
\end{equation*}

\end{theorem}
This result shows that the covering number of $\mathcal{P}$ can be controlled by that of $\mathcal{F}$ in the Lipschitz-continuity condition. Especially, when $0<\lambda<1$, there is no significant difference between the complexities of the two function classes.

In many generalization bounds for ERM, the covering number is employed by letting $d$ be a distribution-dependent metric. For example, in Theorem 2.3 of \citet{mendelson2003few}, the metric $d$ is set to be the norm $\ell_1({\bf Z}\cup{\bf Z}')$ and the generalization bound for ERM is obtained by incorporating the expectation of the covering number, {\it i.e.,} $\mathbb{E}\mathcal{N}(\mathcal{F},\xi,\ell_1({\bf Z}\cup{\bf Z}'))$. Furthermore, the uniform entropy number (UEN) is introduced to eliminate its distribution-dependence \citep[see][]{mendelson2003few}:
\begin{equation}\label{eq:uen}
\mathcal{N}_1(\mathcal{F},\xi,2N) = \max_{{\bf Z},{\bf Z}'} \mathcal{N}(\mathcal{F},\xi,\ell_1({\bf Z}\cup{\bf Z}')).
\end{equation}
{In a similar way, we present two types of UENs for the difference function class $\mathcal{P}$. The first one is expressed as 
\begin{align}\label{eq:uen1}
&\mathcal{N}_1(\mathcal{P},\xi,N):=\max_{{\bf S}^\dag\subset \mathcal{Z}\times\mathcal{Z} } \mathcal{N}(\mathcal{P},\xi,\ell_1({\bf S}^\dag)),
\end{align}
which is of a standard UEN form. When the event $\mathcal{E}_1$ occurs, the second one is defined based on the positional relationship between samples and their ghosts:
\begin{align}\label{eq:uen2}
\mathcal{N}_1^r(\mathcal{P},\xi,N):= \max_{\mathcal{C}_r(\mathcal{Z})}\, \max_{\{C_n\}\subset\mathcal{C}_r(\mathcal{Z})}\, \max_{\{{\bf s}_{[n]}\in C_n\times C_n\}}  \mathcal{N}(\mathcal{P},\xi,\ell_1({\bf S}^\dag)),
\end{align}
where $\max_{\mathcal{C}_r(\mathcal{Z})}$ is taken over all possibilities of $r$-radius cover of $\mathcal{Z}$ w.r.t.~the Euclidean norm; $\max_{\{C_1,\cdots,C_N\}\subset \mathcal{C}_\xi}$ is achieved by selecting $N$ elements $C_1,\cdots,C_N$ from the selected $\mathcal{C}_r(\mathcal{Z})$; and $\max_{\{{\bf s}_{[n]}\in C_n\times C_n\}} $ is achieved by selecting the points ${\bf s}_{[n]}$ from the corresponding sets $C_n\times C_n$ ($1\leq n\leq N$).} Then, we arrive at a distribution-free upper bound of $\mathbb{E}\mathcal{N}(\mathcal{P},\xi,\ell_1({\bf S}^\dag))$:
\begin{lemma}\label{lem:uen}
Given a sample set ${\bf Z}=\{{\bf z}_{1},\cdots,{\bf z}_{N}\}$ and its vicinity ghost ${\bf Z}^\dag:=\{{\bf z}'_{\pi_*(1)},\cdots,{\bf z}'_{\pi_*(N)}\}$, let ${\bf S}^\dag = \{{\bf s}_{[1]},\cdots,{\bf s}_{[N]}\}$ with ${\bf s}_{[n]}=({\bf z}'_{\pi_*(n)},{\bf z}_n)$ ($1\leq n\leq N$). For any function class $\mathcal{P}\subset \mathbb{R}^{\mathcal{Z}\times \mathcal{Z}}$, it holds that
\begin{align}\label{eq:uen3}
\mathbb{E}\mathcal{N}(\mathcal{P},\xi,\ell_1({\bf S}^\dag))
\leq  \mathcal{N}_1^r(\mathcal{P},\xi,N) + c \cdot {\rm e}^{\frac{-Nr^2}{2 K}} \cdot \mathcal{N}_1(\mathcal{P},\xi,N),
\end{align}
where $c>0$ is an absolute constant.
\end{lemma}

{\bf Proof:} It follows from Theorem \ref{thm:point} that
\begin{align}\label{eq:uen3}
\mathbb{E}\mathcal{N}(\mathcal{P},\xi,\ell_1({\bf S}^\dag))
=& \mathbb{E}\big\{\mathcal{N}(\mathcal{P},\xi,\ell_1({\bf S}^\dag))\,|\, \mathcal{E}_1\big\}\cdot \mathbb{P}\{\mathcal{E}_1 \}+\mathbb{E}\big\{\mathcal{N}(\mathcal{P},\xi,\ell_1({\bf S}^\dag))\,|\, \mathcal{E}_2\big\}\cdot \mathbb{P}\{\mathcal{E}_2 \}\nonumber\\
\leq & \mathbb{E}\big\{\mathcal{N}(\mathcal{P},\xi,\ell_1({\bf S}^\dag))\,|\, \mathcal{E}_1\big\} + c \cdot {\rm e}^{\frac{-Nr^2}{2 K}}\mathbb{E}\big\{\mathcal{N}(\mathcal{P},\xi,\ell_1({\bf S}^\dag))\,|\, \mathcal{E}_2\big\}\nonumber\\
\leq & \mathcal{N}_1^r(\mathcal{P},\xi,N) + c \cdot {\rm e}^{\frac{-Nr^2}{2 K}}\cdot \mathcal{N}_1(\mathcal{P},\xi,N).
\end{align}
This completes the proof.\hfill$\blacksquare$

As shown in Lemma \ref{lem:uen}, the term $\mathbb{E}\mathcal{N}(\mathcal{P},\xi,\ell_1({\bf S}^\dag))$ can be bounded by the sum of two terms: $\mathcal{N}_1^r(\mathcal{P},\xi,N)$ that measures the complexity of $\mathcal{P}$ when the event $\mathcal{E}_1$ occurs, and $\mathcal{N}_1(\mathcal{P},\xi,N)$ defined in the general UEN form whose weight decays to {\it zero} exponentially as the sample size $N$ approaches {\it infinity}. Therefore, the term $\mathbb{E}\mathcal{N}(\mathcal{P},\xi,\ell_1({\bf S}^\dag))$ is dominated by $\mathcal{N}_1^r(\mathcal{P},\xi,N)$ when $N$ is large. This result indicates that the positional relationship between samples and their ghost is crucial to analyzing VRM's generalization performance.



\section{Main Results}\label{sec:main}

In this section, we present the generalization bounds for VRM and then analyze the behavior of the quantity $\Omega_\nu$ that measures the effectiveness of introducing vicinal functions.

\subsection{Generalization Bounds for VRM}

{Let $\Lambda^*_\mathcal{P}(\xi/4,\ell_1({\bf S}^{\dag}))$ be the {cover} that achieves the covering number $\mathcal{N}(\mathcal{P},\xi/4,\ell_1({\bf S}^\dag))$, and denote
{\begin{align}\label{eq:U}
\Omega_\nu :=\sup_{p\in \Lambda^*_\mathcal{P}(\xi/4,\ell_1({\bf S}^{\dag})) }\mathbb{E}\Big\{\frac{1}{N}\sum_{n=1}^Np({\bf s}_{[n]})\Big\}.
\end{align}}
Then, we obtain a distribution-dependent generalization bound for VRM, which is based on the expectation of covering numbers. 

\begin{theorem}\label{thm:bound1}
Assume that F is a function class with the range $[a, b]$. For any $\xi>0$ and $N\geq\frac{8(b-a)^2}{\xi^2}$, there holds that with probability at least $1-t$,
\begin{align}\label{eq:vrm.bound1}
\sup_{f\in\mathcal{F}}\{\mathrm{R}f-\mathrm{R}_\nu f\} \leq 4\Omega_\nu+\sqrt{ \frac{32(b-a)\big(\log \mathbb{E}\mathcal{N}(\mathcal{P},\xi/4,\ell_1({\bf S}^\dag)) - \log (t/2)\big)}{N}  }.
\end{align}
\end{theorem}}
If the quantity $\Omega_\nu$ is ignored, this bound has the same form as that of the classical generalization bound \cite[see][Theorem 2.3]{mendelson2003few} except for the different constants. To eliminate the distribution dependence, we then use Lemma \ref{lem:uen} to obtain a UEN-based generalization bound that is distribution-free.
\begin{corollary}\label{cor:bound2}
For any $N\geq\frac{8(b-a)^2}{\xi^2}$, then there holds that with probability at least $1-t$,
\begin{align}\label{eq:vrm.bound}
\sup_{f\in\mathcal{F}}\{\mathrm{R}f-\mathrm{R}_\nu f\} \leq 4\Omega_\nu
&+ \sqrt{\frac{32(b-a)^2}{N}} \Big[\log\Big(\mathcal{N}_1^r(\mathcal{P},\xi,N) \nonumber\\
&+   c \cdot {\rm e}^{\frac{-Nr^2}{2 K}}\mathcal{N}_1(\mathcal{P},\xi,N) \Big)- \log (t/2)\Big]^{\frac{1}{2}},
\end{align}
where $c>0$ is an absolute constant.
\end{corollary}

The two bounds show that VRM's generalization performance is influenced by two factors: the complexity of $\mathcal{P}$ and the quantity $\Omega_\nu$. As addressed in Theorem \ref{thm:covering}, since the complexity of $\mathcal{P}$ can be controlled by that of the original function class $\mathcal{F}$, the quantity $\Omega_\nu$ plays an essential role in analyzing VRM's generalization performance. If the quantity $\Omega_\nu$ is negative, the bounds \eqref{eq:vrm.bound1} and \eqref{eq:vrm.bound} will become sharper than those for ERM, {\it i.e.,} VRM has a better generalization performance than ERM. In contrast, a positive $\Omega_\nu$ means that the choice of ${\rm V}$ is not suitable, while the bounds can also be tight if $\mathcal{F}$ has a low complexity. These findings are in accordance with Vapnik's original arguments about VRM \citep[see][]{vapnik1999nature}.

\subsection{Analysis of $\Omega_\nu$}\label{sec:omega}

Here, we give a detailed discussion on the quantity $\Omega_\nu$. The following lemma provides a chance to analyze the behavior of $\Omega_\nu$ based on the original function class $\mathcal{F}$ rather than the difference function class $\mathcal{P}$.

\begin{proposition}\label{pro:analysis}
{If $\phi$ satisfies the $\lambda$-Lipschitz condition that $|\phi(f,{\bf z}) - \phi(h,{\bf z})|\leq \lambda |f({\bf z})-h({\bf z})|$, $\forall f,h\in\mathcal{F}$, $\lambda\in(0,1)$}, then there exists an at most countable set $\Lambda^*\subset \mathcal{F}$ such that
\begin{align}
\Omega_\nu = & \sup_{f\in \Lambda^* }\Big\{ \int f({\bf z})\;  d\,\big[ {\rm P}({\bf z}) - \mathbb{E}_{{\bf a}\sim D({\cal Z})} \big\{ {\rm V}\big({\bf z}|\theta({\bf a})\big)\big\}  \big]\Big\}\label{eq:analysis2}
\\%
=&\sup_{f\in \Lambda^* }\Big\{ \mathrm{R}f - \mathrm{R} \phi (f,{\bf z})\Big\}.\label{eq:analysis1}
\end{align}
\end{proposition}
Compared with the quantity $\sup_{f\in\mathcal{F}}\big\{\mathrm{R}f-\mathrm{R}
\phi(f,{\bf z})\big\}$ in the right-hand side of \eqref{eq:rhs1}, the quantity $\Omega_\nu$ only takes the maximum over a countable subset $\Lambda^*$ rather than the supremum over the whole function class $\mathcal{F}$. It provides the convenience for studying the learning problems with high-complexity function classes. This lemma also suggests that the behavior of $\Omega_\nu$ should be analyzed from two different perspectives: 1) the approximation to the underlying distribution ${\rm P}({\bf z})$ [see (V1)]; and 2) the modification to the function $f$ by convolving it with the vicinal function ${\rm V}$ [see (V2)].

Let $\widetilde{f}\in\Lambda^*$ be the function that achieves the maximization operation in Proposition \ref{pro:analysis}. Recalling \eqref{eq:analysis2}, since the value of $\Omega_\nu$ is also determined by the difference between the distributions ${\rm P} ({\bf z})$ and $\mathbb{E}_{{\bf a}\sim D({\cal Z})} \big\{ {\rm V}[{\bf z}|\theta({\bf a})]\big\}$, a negative $\Omega_\nu$ means that the function $\widetilde{f}$ causes a larger error on the distribution $\mathbb{E}_{{\bf a}\sim D({\cal Z})} \big\{ {\rm V}\big({\bf z}|\theta({\bf a})\big)\big\}$ than on the distribution ${\rm P}({\bf z})$. Alternatively, as shown in \eqref{eq:analysis1}, the negative value of $\Omega_\nu$ means that the modified function $\phi (\widetilde{f},{\bf z})$ causes a larger expected risk than the original function $\widetilde{f}$, {\it i.e.,} a reasonable vicinity function should amplify the error caused by $\widetilde{f}$ in order to alleviate the overfitting probably appearing in ERM.

Moreover, let $f_\nu$ be the learning result by using an algorithm to minimize the vicinal risk $\mathrm{R}_\nu f$ over $\mathcal{F}$. Denote
\begin{align}\label{eq:pi}
\eta_\nu:=& \int f_\nu({\bf z})\;  d\,\big[ {\rm P}({\bf z}) - \mathbb{E}_{{\bf a}\sim D({\cal Z})} \big\{ {\rm V}\big({\bf z}|\theta({\bf a})\big)\big\}  \big]\nonumber\\
\eta^{(1)}_\nu:= & \int f_\nu({\bf z})\;  d\,\big[ {\rm P}({\bf z}) - \mathrm{P}_\nu ({\bf z})   \big]\nonumber\\
\eta^{(2)}_\nu:= &\int f_\nu({\bf z})\;  d\,\big[   \mathbb{E}_{{\bf a}\sim D({\cal Z})} \big\{ {\rm V}\big({\bf z}|\theta({\bf a})\big)\big\} -\mathrm{P}_\nu ({\bf z})   \big],
\end{align}
and it holds that $\eta_\nu = \eta^{(1)}_\nu-\eta^{(2)}_\nu$. The term $\eta^{(1)}_\nu$ describes the discrepancy between the underlying distribution $\mathrm{P}({\bf z}) $ and the vicinal distribution $\mathrm{P}_\nu ({\bf z}) $. It always holds that $\eta^{(1)}_\nu\geq0$. Since $\mathbb{E}_{{\bf Z}\sim (D(\mathcal{Z}))^N} \{\mathrm{P}_\nu ({\bf z}) \}= \mathbb{E}_{{\bf a}\sim D({\cal Z})} \big\{ {\rm V}[{\bf z}|\theta({\bf a})]\big\}$, the vicinal distribution $\mathrm{P}_\nu ({\bf z})$ is an empirical estimate of $\mathbb{E}_{{\bf a}\sim D({\cal Z})} \big\{ {\rm V}[{\bf z}|\theta({\bf a})]\big\}$. The term $\eta^{(2)}_\nu$ can also be regarded as the function w.r.t. the random variables ${\bf z}_1,{\bf z}_2,\cdots,{\bf z}_N$ with $\mathbb{E}_{{\bf Z}\sim (D(\mathcal{Z}))^N} \{ \eta^{(2)}_\nu \} =0$.

Below, we would like to obtain the lower bound of the probability that $\eta_\nu<0$ to show that the introduction of vicinal function can improve the generalization performance of ERM in a high probability. The validity of $\eta_\nu<0$ requires that $\eta^{(2)}_\nu>\eta^{(1)}_\nu$. Since $\eta^{(2)}_\nu$ is a centered random variable, we will employ the techniques of small ball probability to achieve the lower bound of the probability that $\eta_\nu < 0$. Refer to \citet{li2001gaussian} for more details on small value probability.

\begin{theorem}\label{thm:small}
Assume that there exists a constant $\tau\in\mathbb{R}$ such that $\eta^{(1)}_\nu\leq \tau$ holds for any realization of samples ${\bf z}_1,\cdots,{\bf z}_N$. If there exist two constant $b_1,b_2>0$ such that 
\begin{equation}\label{eq:small.condition}
  \mathbb{E} \mathrm{e}^{-t \eta_\nu^{(2)}} \leq b_1 {\rm e}^{-b_2 t^{\frac{\alpha}{1+\alpha}}},\quad \forall \, \alpha ,t>0,
\end{equation}
then it holds that
\begin{equation}\label{eq:small.result}
\mathbb{P}\{\eta_\nu<0 \} \geq 1-c_1 {\rm e }^{-c_2 \tau^{-\alpha}},
\end{equation}
where $c_1,c_2>0$ are absolute constants.
\end{theorem}
This theorem interprets that under the condition \eqref{eq:small.condition}, the inequality $\eta_\nu <0$ holds with probability at least $1-c_1 {\rm e }^{-c_2 \tau^{-\alpha}}$. To increase the probability that $\eta_\nu<0$, it is necessary to decrease the value of $\tau$, which corresponds to the approximation performance of the vicinal distribution $\mathrm{P}_\nu ({\bf z})$. Namely, if $\mathrm{P}_\nu ({\bf z})$ approximates $\mathrm{P} ({\bf z})$ precisely, the inequality $\eta_\nu<0$ will hold with a high probability. This findings supports the validity of Viewpoint V1 (see Section \ref{sec:intr}), and can be used to examine the feasibility of VRM models based on uniform distributions and Gaussian distributions, proposed by \citet{vapnik1999nature}, respectively.

\begin{remark}\label{rem:vrm2}

Recalling \eqref{eq:mixup}, the mixup vicinal functions actually generate new samples in the way of weighted linear combination of samples, where weights obey a beta distribution. Since the two parameters of the beta distribution share the same value, the new samples tend to lie in the gap among the original samples \citep[see][Fig. 1]{zhang2017mixup}. In this manner, if the sample size is large, the mixup vicinal function can be deemed as the one defined on the vicinities of  sample points, and the corresponding mixup vicinal distribution can approximate $\mathrm{P}({\bf z})$ accurately if the parameter $\alpha$ of ${\rm Beta}(\alpha,\alpha)$ is selected suitably. Therefore, the mixup-based VRM outperforms ERM.

\end{remark}

\section{Conclusion}

In this paper, we study the generalization bounds for VRM and then analyze the factors influencing VRM's generalization
performance. Our results support Vapnik's original arguments
on VRM \citep[see][]{vapnik1999nature} but also provide deeper insights into it. Specifically, we first present the symmetrization inequality
for VRM as the starting point to obtain the generalization
bounds. Compared to the classical symmetrization result
for ERM \citep[see][Lemma 2]{bousquet2004introduction}, the resulting symmetrization inequality for VRM has
some specific considerations. Since the expectation $\mathbb{E} \{\widehat{\mathrm{R}}'f-\mathrm{R}_\nu f\}$ usually is not equal
to {\it zero}, the classical Hoeffding's inequality cannot be directly applied to the VRM setting. Instead, we apply the one-sided concentration inequality for random variables with non-zero means to achieve the upper bound of $\mathbb{P}\{\widehat{\mathrm{R}}'f-\mathrm{R}_\nu f>\frac{\xi}{2}\}$.

It is noteworthy that an arbitrary permutation of ghost samples ${\bf z}'_1,\cdots,{\bf z}'_N$ holds the equality (13), and VRM's inherent characteristics require us to consider the sample pair ${\bf s}_n:=({\bf z}'_{\pi_*(n)},{\bf z}_n)$ ($1\leq n\leq N$) containing a sample ${\bf z}_n$ and its vicinity ghost sample ${\bf z}'_{\pi_*(n)}$ located in the vicinity of sample ${\bf z}_{[n]}$. Therefore, we introduce the concept of vicinity ghost samples to capture the positional relationship between the samples and their ghosts. Moreover, we provide a theoretical guarantee of the existence of the sample pair: the probability that the Euclidean distance between the sample and its vicinity ghost is large will exponentially decay to zero when the sample size $N$ goes to infinity.

Based on the sample pairs, we introduce the difference function class $\mathcal{P}:=\{f({\bf z})-\phi(f,{\bf z}):f\in\mathcal{F}\}$ and then prove that the covering number of $\mathcal{P}$ can be controlled by that of $\mathcal{F}$. To eliminate distribution-dependence, we bound $\mathbb{E}\mathcal{N}(\mathcal{P},\xi,\ell_1({\bf S}^\dag))$ using the sum of two types of UENs for $\mathcal{P}$: the first by selecting $N$ points from $\mathcal{Z}\times\mathcal{Z}$ and the second by selecting elements from a $\xi$-rectangle cover of $\mathcal{Z}\times\mathcal{Z}$. We highlight that the complexity of $\mathcal{P}$ is dominated by the latter UEN when sample $N$ is large, which also implies that the positional relationship between samples and their ghosts plays an important role when studying VRM's properties.

The resulting generalization bounds for VRM show that two factors influence VRM's generalization performance: the complexity of $\mathcal{P}$ and the quantity $\Omega_\nu$. Since the complexity of $\mathcal{P}$ can be controlled by that of $\mathcal{F}$, the quantity $\Omega_\nu$ is crucial to analyzing VRM's generalization performance. We then give a detailed analysis of $\Omega_\nu$ and then provide a strategy to examine whether the choice of vicinal functions is suitable for the learning setting. Finally, we provide a theoretical explanation of some existing VRM models including uniform distribution-based models, Gaussian distribution-based models and mixup-based models.

As addressed in Section \ref{sec:motivation}, one motivation of this paper is to analyze behavior of $\sup_f\{\mathrm{R}f-\mathrm{R}\phi(f,{\bf z})\}$ on a countable subset of $\mathcal{F}$ rather than the whole function class $\mathcal{F}$. Although the inequality \eqref{eq:rhs1} provides a direct way to use the classical results on statistical learning theory for obtaining the generalization bounds for VRM based on several kinds of function class complexities ({\it e.g.} Rademacher complexities), one has to analyze the quantity $\sup_{f\in\mathcal{F}}\big\{\mathrm{R}f-\mathrm{R}\phi(f,{\bf z})\big\}$, which is contrary to the motivation. It is still challenging to develop the appropriate concentration inequalities for obtain the Rademacher complexity-based bounds for VRM. Our future works will focus on this issue as well as the complexity analysis for the function classes $\{\phi(f,{\bf z}): f\in\mathcal{F}\}$ generated from different vicinity functions.



\appendix


\section{Useful Inequalities}\label{sup:sym}

In this section, we present two useful inequalities as the starting point for obtaining generalization bounds for VRM.

\subsection{Symmetrization Inequalities}

In order to achieve the upper bound of \eqref{eq:def.bound}, the first step is to apply the symmetrization inequality to replace the expected risk $\mathrm{R}f$ with the empirical risk $\widehat{\mathrm{R}}'f$ computed on the ghost sample set ${\bf Z}' = \{{\bf z}'_1,\cdots,{\bf z}'_N\}$, where ${\bf z}_n$ and ${\bf z}'_n$ are independent and have the same distribution.

\begin{lemma}\label{lem:sym}
Assume that $\mathcal{F}$ is a function class with the range $[a,b]$. Let ${\bf Z}$ and $\mathbf{Z}'$ be an i.i.d sample set drawn from $\mathcal{Z}$ and its ghost, respectively. Then, there holds that for any $\xi>0$ and $N\geq\frac{8(b-a)^2}{\xi^2}$,
\begin{align}\label{eq:sym.vrm}
    \mathbb{P}\Big\{\sup_{f\in\mathcal{F}}\{\mathrm{R}f-\mathrm{R}_\nu f\}>\xi\Big\}
   \leq 2\mathbb{P}\Big\{\sup_{f\in\mathcal{F}}\{\widehat{\mathrm{R}}'f-\mathrm{R}_\nu f\}>\frac{\xi}{2}\Big\}.
\end{align}
\end{lemma}
This result is similar with the symmetrization inequality for ERM \citep[see][Lemma 2]{bousquet2004introduction}, so we omit its proof. Since the intractable expected risk $\mathrm{R}f$ is replaced with the empirical risk $\widehat{\mathrm{R}}'f$ computed on the ghost samples, we consider the following issues: 1) how to obtain the upper bound of $\mathbb{P}\{\{\widehat{\mathrm{R}}'f-\mathrm{R}_\nu f\}>\frac{\xi}{2}\}$; and 2) how to convert the supremum operation $\sup_{f\in\mathcal{F}}$ into the summation operation $\sum_{h\in \Lambda}$, where $\Lambda$ is a subset of $\mathcal{F}$ with some special structure, {\it e.g.,} the cover of $\mathcal{F}$. Next, we present the specific concentration inequality to handle the first issue.


\subsection{Concentration Inequalities}

In order to achieve the generalization bound of a learning process, one generally needs to find the appropriate concentration inequalities. In the ERM setting, the concentration inequalities, holding for the random variables with zero means, describes the probability that the discrepancy between the empirical risk $\widehat{\mathrm{R}}f$ and the expected risk $\mathrm{R}f$ is larger than a positive constant. In contrast to the ERM setting, since the exception $\mathbb{E}\{ \widehat{\mathrm{R}}'f-\mathrm{R}_\nu f \}$ usually is not equal to {\it zero}, the classical concentration inequalities ({\it e.g.} Hoeffding's inequality) cannot be directly applied to the VRM setting. Instead, we will use the one-sided concentration inequality for random variables with non-zero means to bound $\mathbb{P}\{\widehat{\mathrm{R}}'f-\mathrm{R}_\nu f>\frac{\xi}{2}\}$:


%
\begin{lemma}\label{lem:hoeffding}
Let $x_1,\cdots,x_N$ be independent random variables with $a_n\leq x_n - \mathbb{E} x_n \leq b_n$ almost surely for any $1\leq n\leq N$. Then, it holds that for any $\xi>0$,
\begin{align}\label{eq:hoeffding}
\mathbb{P}\Big\{ \sum_{n=1}^N x_n>\xi  \Big\}\leq\exp\Big( \frac{-2\big[\xi-\sum_{n=1}^N \mathbb{E}x_n\big]^2 }{\sum_{n=1}^N (b_n-a_n)^2 }  \Big).
\end{align}
\end{lemma}
This one-sided inequality for random variables with non-zero means is a variation of Hoeffding's inequality and especially, if $\mathbb{E}x_n=0$ holds for any $1\leq n\leq N$, the proposed inequality \eqref{eq:hoeffding} coincides with Hoeffding's inequality.



\section{Proofs of Main Results}\label{sup:proof}

In this section, we prove Theorem \ref{thm:point}, Theorem \ref{thm:covering}, Theorem \ref{thm:bound1} and Proposition \ref{pro:analysis}, respectively.

\subsection{Proof of Theorem \ref{thm:point}}

In order to prove Theorem \ref{thm:point}, we first need to introduce the Dvoretzky-Kiefer-Wolfowitz (DKW) inequality for random vectors \citep[see][]{kiefer1958deviations,kiefer1961large}:


{
\begin{lemma}[DKW Inequality]\label{lem:DKW}
Let $z$ be a random variable defined on $\mathbb{R}$ with the cdf ${\rm F}(t)$, $t\in\mathbb{R}$. Then, there exists a positive constant $c$ (not depending on ${\rm F}$) such that for any $\xi>0$,
\begin{equation*}
\mathbb{P}\left\{ \sup_{t\in\mathbb{R}} |\mathrm{F}(t)-\widehat{\mathrm{F}}(t)| > \xi   \right\} \leq c\cdot {\rm e}^{-2N\xi^2}.
\end{equation*}
%
\end{lemma}

Now, we come up with the proof of Theorem \ref{thm:point}.

{\bf Proof of Theorem \ref{thm:point}:} According to \eqref{eq:ecdf}, since $(a-b)^2 \leq 2\big((a-c)^2+(b-c)^2\big)$ holds for any $a,b,c\in\mathbb{R}$, we have 
\begin{align}\label{eq:ecdf.pr1}
&\max_{n\in\{1,\cdots,N\}}\, d^2_{\widehat{\mathrm{F}}}({\bf z}_n, {\bf z}_{\pi_*(n)})\nonumber\\
 =&  \max_{n\in\{1,\cdots,N\}}\, \sum_{k=1}^K \big(\widehat{\mathrm{F}}_k ({\bf z}_n) -\widehat{\mathrm{F}}_k ({\bf z}_{\pi_*(n)})\big)^2\nonumber\\
\leq & K\max_{n\in\{1,\cdots,N\}} \max_{k\in\{1,\cdots,K\}} \big(\widehat{\mathrm{F}}_k ({\bf z}_n) -\widehat{\mathrm{F}}_k ({\bf z}_{\pi_*(n)})\big)^2\nonumber\\
\leq & 2K\max_{n\in\{1,\cdots,N\}} \max_{k\in\{1,\cdots,K\}}\left[ \big(\mathrm{F}_k ({\bf z})-\widehat{\mathrm{F}}_k ({\bf z}_n)\big)^2+ \big(\mathrm{F}_k ({\bf z}) -\widehat{\mathrm{F}}_k ({\bf z}_{\pi_*(n)})\big)^2\right].
\end{align}
Therefore, since ${\bf Z}$ and ${\bf Z}'$ have the same distribution, it holds that for any $\xi>0$, 
\begin{align}\label{eq:ecdf.pr2}
&\mathbb{P}\left\{ \max_{n\in\{1,\cdots,N\}}\, d_{\widehat{\mathrm{F}}}({\bf z}_n, {\bf z}_{\pi_*(n)})>\xi\right\}\nonumber\\
=&\mathbb{P}\left\{ \max_{n\in\{1,\cdots,N\}}\, d^2_{\widehat{\mathrm{F}}}({\bf z}_n, {\bf z}_{\pi_*(n)})>\xi^2\right\}\nonumber\\
\leq & \mathbb{P}\left\{  \max_{n\in\{1,\cdots,N\}} \max_{k\in\{1,\cdots,K\}}\left[ \big(\mathrm{F}_k ({\bf z})-\widehat{\mathrm{F}}_k ({\bf z}_n)\big)^2+ \big(\mathrm{F}_k ({\bf z}) -\widehat{\mathrm{F}}_k ({\bf z}_{\pi_*(n)})\big)^2\right]> \frac{\xi^2}{2K} \right\}\nonumber\\
\leq & \mathbb{P}\left\{  \max_{n\in\{1,\cdots,N\}} \max_{k\in\{1,\cdots,K\}}\left[ \big(\mathrm{F}_k ({\bf z})-\widehat{\mathrm{F}}_k ({\bf z}_n)\big)^2\right]> \frac{\xi^2}{4K} \right\}\nonumber\\
= & \mathbb{P}\left\{  \max_{n\in\{1,\cdots,N\}} \max_{k\in\{1,\cdots,K\}}\left[\big|\mathrm{F}_k ({\bf z})-\widehat{\mathrm{F}}_k ({\bf z}_n)\big|\right]> \frac{\xi}{2\sqrt{K}} \right\}.
\end{align}
The combination of Lemma \ref{lem:DKW} and \eqref{eq:ecdf.pr2} leads to
\begin{align*}
&\mathbb{P}\left\{ \max_{n\in\{1,\cdots,N\}}\, d_{\widehat{\mathrm{F}}}({\bf z}_n, {\bf z}_{\pi_*(n)})>\xi\right\}
\leq  c \cdot {\rm e}^{\frac{-N\xi^2}{2 K}},\quad \forall\, \xi>0,
\end{align*}
where $c>0$ is an absolute constant. This completes the proof.\hfill$\blacksquare$
}
%

\subsection{Proof of Theorem \ref{thm:covering}}

{\bf Proof of Theorem \ref{thm:covering}:} Given $\xi>0$, let $\Lambda$ and $\Lambda'$ be the covers of $\mathcal{F}$ that achieves the covering numbers $\mathcal{N}\big(\mathcal{F},\frac{\xi}{2+\lambda},\ell_1({\bf Z})\big)$ and $\mathcal{N}\big(\mathcal{F},\frac{\xi}{2+\lambda},\ell_1({\bf Z}')\big)$ respectively, {\it i.e.,} $|\Lambda| = \mathcal{N}\big(\mathcal{F},\frac{\xi}{2+\lambda},\ell_1({\bf Z})\big)$ and $|\Lambda'| = \mathcal{N}\big(\mathcal{F},\frac{\xi}{2+\lambda},\ell_1({\bf Z}')\big)$. Therefore, for any $f\in\mathcal{F}$, there exist $h\in\Lambda$ and $h'\in\Lambda'$ such that
\begin{equation*}
\frac{1}{N} \sum_{n=1}^N |f({\bf z}_n) - h({\bf z}_n)| \leq \frac{\xi}{2+\lambda},
\end{equation*}
and
\begin{equation*}
\frac{1}{N} \sum_{n=1}^N |f({\bf z}'_n) - h'({\bf z}'_n)| \leq \frac{\xi}{2+\lambda}.
\end{equation*}
Then, we have
\begin{align*}
&\frac{1}{N} \sum_{n=1}^N \big| [f({\bf z}'_{\pi_*(n)}) - \phi(f,{\bf z}_n)] -  [h({\bf z}'_{\pi_*(n)}) - \phi(h,{\bf z}_n)]  \big| \nonumber\\
\leq & \frac{1}{N} \sum_{n=1}^N \big| f({\bf z}'_{\pi_*(n)}) -h({\bf z}'_{\pi_*(n)})\big|+\frac{1}{N} \sum_{n=1}^N\big| \phi(f,{\bf z}_n)- \phi(h,{\bf z}_n)  \big|\nonumber\\
=& \frac{1}{N} \sum_{n=1}^N \big| f({\bf z}'_{\pi_*(n)}) -h({\bf z}'_{\pi_*(n)})\big|+\frac{1}{N} \sum_{n=1}^N\big| \phi(f,{\bf z}_n)- \phi(h,{\bf z}_n)  \big|\nonumber\\
\leq &  \frac{1}{N} \sum_{n=1}^N \big| f({\bf z}'_{\pi_*(n)}) -h({\bf z}'_{\pi_*(n)})\big|+\frac{\lambda}{N} \sum_{n=1}^N \big| f({\bf z}_n) -h({\bf z}_n)\big|\nonumber\\
\leq &  \frac{1}{N} \sum_{n=1}^N \big| f({\bf z}'_{\pi_*(n)}) -h'({\bf z}'_{\pi_*(n)})\big|+ \frac{1}{N} \sum_{n=1}^N \big| h({\bf z}'_{\pi_*(n)}) -h'({\bf z}'_{\pi_*(n)})\big|+\frac{\lambda}{N} \sum_{n=1}^N \big| f({\bf z}_n) -h({\bf z}_n)\big|\nonumber\\
\leq & \frac{\xi}{2+\lambda}+\frac{\xi}{2+\lambda}+\frac{\xi\cdot\lambda}{2+\lambda}=\xi.
\end{align*}
which implies that $\{h({\bf z}_1)-\phi(h,{\bf z}_2)\}_{h\in\Lambda}$ is a $\xi$-radius cover of $\mathcal{P}$ w.r.t. $\ell_1({\bf S}^\dag)$. Therefore, we arrive at
\begin{equation*}
\mathcal{N}\big(\mathcal{P},\xi,\ell_1({\bf S}^\dag) \big)\leq \mathcal{N}\Big(\mathcal{F},\frac{\xi}{2+\lambda},\ell_1({\bf Z}) \Big).
\end{equation*}

On the other hand, if $\Omega$ is a $\xi$-radius cover of $\mathcal{P}$ that achieves the covering number $\mathcal{N}(\mathcal{P},\xi,\ell_1({\bf S}^\dag))$, {\it i.e.,} $|\Omega| = \mathcal{N}(\mathcal{P},\xi,\ell_1({\bf S}^\dag))$. Then, for any $p\in\mathcal{P}$, there exists a $q\in\Omega$ such that
\begin{align*}
\frac{1}{N} \sum_{n=1}^N |p({\bf s}_{[n]}) - q({\bf s}_{[n]})|
= \frac{1}{N} \sum_{n=1}^N \big| [f({\bf z}'_{\pi_*(n)}) - \phi(f,{\bf z}_n)] -  [h({\bf z}'_{\pi_*(n)}) - \phi(h,{\bf z}_n)]  \big|\leq \xi,
\end{align*}
where ${\bf s}_{[n]}=({\bf z}'_{\pi_*(n)},{\bf z}_n)$, $p({\bf s}) = f({\bf z}'_{\pi_*(n)})-\phi(f,{\bf z})$ and $q({\bf s})= h({\bf z}'_{\pi_*(n)})-\phi(h,{\bf z})$.
For any $\lambda\in (0,1)$, we then have
\begin{align*}
& \frac{1-\lambda}{N} \sum_{n=1}^N \big| f({\bf z}_n) -h({\bf z}_n)\big|\nonumber\\
\leq & \frac{1}{N} \sum_{n=1}^N\Big| \big| f({\bf z}'_{\pi_*(n)}) -h({\bf z}'_{\pi_*(n)})\big|-| \phi(f,{\bf z}_n)- \phi(h,{\bf z}_n)  \big|\Big|\nonumber\\
\leq &
\frac{1}{N} \sum_{n=1}^N \big| [f({\bf z}'_{\pi_*(n)}) - \phi(f,{\bf z}_n)] -  [h({\bf z}'_{\pi_*(n)}) - \phi(h,{\bf z}_n)]  \big|\nonumber\\
=& \frac{1}{N} \sum_{n=1}^N \big| p({\bf s}_n) - q({\bf s}_n)  \big| <\xi,
\end{align*}
which implies that the set $\{h\}$ is a $\frac{\xi}{1-\lambda}$-radius cover of $\mathcal{F}$ w.r.t. $\ell_1({\bf Z})$. Therefore, we arrive at
\begin{equation*}
\mathcal{N}\Big(\mathcal{F},\frac{\xi}{1-\lambda},\ell_1({\bf Z}) \Big)\leq \mathcal{N}\big(\mathcal{P},\xi,\ell_1({\bf S}^\dag) \big),\quad \forall \lambda\in(0,1).
\end{equation*}
This completes the proof. \hfill$\blacksquare$

\subsection{Proof of Theorem \ref{thm:bound1}}

{\bf Proof of Theorem \ref{thm:bound1}:} According to Lemma \ref{lem:sym}, given an arbitrary $\xi>0$, we have for any $N\geq\frac{8(b-a)^2}{\xi^2}$,
\begin{align}\label{eq:bound.pr1}
    &\mathbb{P}\left\{\sup_{f\in\mathcal{F}}
  \mathrm{R}f-  \mathrm{R}_\nu f>\xi\right\}\nonumber\\
\leq& 2\mathbb{P}\left\{\sup_{f\in\mathcal{F}}
\widehat{\mathrm{R}}'f-\mathrm{R}_\nu f>\frac{\xi}{2}\right\}\nonumber\\
=&2\mathbb{P}\left\{\sup_{f\in\mathcal{F}}
\frac{1}{N}\sum_{n=1}^{N}\big(f({\bf z}'_n)-\phi(f,{\bf z}_n)\big)>\frac{\xi}{2}\right\}\nonumber\\
=&2\mathbb{P}\left\{\sup_{f\in\mathcal{F}}
\frac{1}{N}\sum_{n=1}^{N}\big(f({\bf z}'_{\pi_*(n)})-\phi(f,{\bf z}_n)\big)>\frac{\xi}{2}\right\}\nonumber\\
=&2\mathbb{P}\left\{\sup_{p\in\mathcal{P}}
\frac{1}{N}\sum_{n=1}^{N}p({\bf s}_{[n]})>\frac{\xi}{2}\right\},
\end{align}
where ${\bf s}_{[n]} = ({\bf z}'_{\pi_*(n)},{\bf z}_{n})$.

Assume that  $p_o$ is the function that achieves the supremum \eqref{eq:bound.pr1}, that is,
\begin{equation*}
\frac{1}{N}\sum_{n=1}^{N}p_o({\bf s}_{[n]})>\frac{\xi}{2}.
\end{equation*}
Fix a realization of ${\bf Z}$ and ${\bf Z}'$ and denote ${\bf S}^{\dag}=\{ {\bf s}_{[1]},\cdots, {\bf s}_{[N]}\}$ with ${\bf s}_{[n]} = ({\bf z}'_{\pi_*(n)},{\bf z}_{n})$. Let $\Lambda_\mathcal{P}(\xi/4,\ell_1({\bf S}^{\dag}))$ be a $\xi/4$-radius cover of $\mathcal{P}$ w.r.t. the $\ell_1({\bf S}^{\dag})$ norm. Then, there must be a $q_o\in\Lambda_\mathcal{P}(\xi/4,\ell_1({\bf S}^{\dag}))$ such that
\begin{equation*}
   \frac{1}{N}\sum_{n=1}^N\left|p_o({\bf s}_{[n]})-q_o({\bf s}_{[n]})\right|  < \frac{\xi}{4}.
\end{equation*}
Since
\begin{equation*}
\frac{1}{N}\sum_{n=1}^N\big(p_o({\bf s}_{[n]})-q_o({\bf s}_{[n]})\big)\leq
   \frac{1}{N}\sum_{n=1}^N\left|p_o({\bf s}_{[n]})-q_o({\bf s}_{[n]})\right|,
\end{equation*}
it is followed from the triangle inequality that
\begin{equation*}
   \frac{1}{N}\sum_{n=1}^Nq_o({\bf s}_{[n]})>\frac{\xi}{4}.
\end{equation*}
Furthermore, let $\Lambda^*_\mathcal{P}(\xi/4,\ell_1({\bf S}^{\dag}))$ be the cover achieving the covering number $\mathcal{N}\big(\mathcal{P},\xi/4,\ell_1({\bf S}^{\dag})\big)$. Since the range of $\mathcal{P}$ is $[a-b,b-a]$, we have
\begin{align}\label{eq:bas2.M}
&\mathbb{P}\Big\{\sup_{f\in\mathcal{F}}
\frac{1}{N}\sum_{n=1}^{N}\big(f({\bf z}'_{\pi_*(n)})-\phi(f,{\bf z}_n)\big)>\frac{\xi}{2}\Big\}\nonumber\\
=&\mathbb{P}\Big\{\sup_{p\in\mathcal{P}}\frac{1}{N}\sum_{n=1}^Np({\bf s}_{[n]})>\frac{\xi}{4}\Big\}\\
\leq&\mathbb{P}\Big\{\sup_{q\in \Lambda^*_\mathcal{P}(\xi/4,\ell_1({\bf S}^{\dag}))}
\frac{1}{N}\sum_{n=1}^Nq({\bf s}_{[n]})>\frac{\xi}{4}\Big\}\nonumber\\
\leq&\mathbb{E}\mathcal{N}\big(\mathcal{P},\xi/4,\ell_1({\bf S}^{\dag})\big)\max_{q\in \Lambda^*_\mathcal{P}(\xi/4,\ell_1({\bf S}^{\dag}))}
\mathbb{P}\Big\{\frac{1}{N}\sum_{n=1}^Nq({\bf s}_{[n]})>\frac{\xi}{4}\Big\}\nonumber\\
\leq& \mathbb{E}\mathcal{N}\big(\mathcal{P},\xi/4,\ell_1({\bf S}^{\dag})\big)\exp\Big( \frac{-\big[N\xi-4\mathbb{E}\{\sum_{n=1}^Nq_1({\bf s}_{[n]})\}\big]^2 }{32N(b-a)^2 }  \Big)\nonumber,
\end{align}
where
\begin{align}
q_1:=& \arg\max_{q\in \Lambda^*_\mathcal{P}(\xi/4,\ell_1({\bf S}^{\dag}))}
\mathbb{P}\Big\{\frac{1}{N}\sum_{n=1}^Nq({\bf s}_{[n]})>\frac{\xi}{4}\Big\}.
\end{align}
Alternatively, with the probability at least $1-t$, there holds that for any $f\in\mathcal{F}$,
\begin{align*}
\mathrm{R}f\leq &\mathrm{R}_\nu f +\frac{4\mathbb{E}\{\sum_{n=1}^Nq_1({\bf s}_n)\}}{N}+ \sqrt{ \frac{32(b-a)\big(\log \mathbb{E}\mathcal{N}(\mathcal{P},\xi/4,\ell_1({\bf S}^\dag)) - \log (t/2)\big)}{N}  }\nonumber \\
\leq &\mathrm{R}_\nu f +\max_{q\in \Lambda^*_\mathcal{P}(\xi/4,\ell_1({\bf S}^{\dag})) }\mathbb{E}\Big\{\frac{4}{N}\sum_{n=1}^Nq({\bf s}_{[n]})\Big\}\nonumber\\
&+ \sqrt{ \frac{32(b-a)\big(\log \mathbb{E}\mathcal{N}(\mathcal{P},\xi/4,\ell_1({\bf S}^\dag)) - \log (t/2)\big)}{N}  }\nonumber
\end{align*}
This completes the proof.\hfill$\blacksquare$

\subsection{Proof of Proposition \ref{pro:analysis}}

{\bf Proof of Proposition \ref{pro:analysis}:} According to Theorem \ref{thm:covering}, if $\Lambda^*_\mathcal{P}(\xi/4,\ell_1({\bf S}^{\dag}))$ is the cover that achieves the covering number $\mathcal{N}(\mathcal{P},\xi/4,\ell_1({\bf S}^\dag))$, there exists another {cover} $\Lambda^*\subset\mathcal{F}$ that achieves the covering number $\mathcal{N}\big(\mathcal{F},\frac{\xi}{4(1-\lambda)},\ell_1({\bf Z}) \big)$ and it holds that $|\Lambda^*| \leq |\Lambda^*_\mathcal{P}(\xi/4,\ell_1({\bf S}^{\dag}))|$. Therefore, the set $\Lambda^*$ is at most countable as well. Then, we have

{\begin{align*}
\Omega_\nu = & \sup_{p\in \Lambda^*_\mathcal{P}(\xi/4,\ell_1({\bf S}^{\dag})) }\mathbb{E}\left\{\frac{1}{N}\sum_{n=1}^Np({\bf s}_{[n]})\right\}\nonumber\\
= &\sup_{f\in \Lambda^* } \mathbb{E}\left\{\frac{1}{N}\sum_{n=1}^N\big(f({\bf z}'_{\pi_*(n)})-\phi (f,{\bf z}_n)\big)\right\}\nonumber\\
=&\sup_{f\in \Lambda^* } \mathbb{E}\left\{\frac{1}{N}\sum_{n=1}^N\big(f({\bf z}'_{n})-\phi (f,{\bf z}_n)\big)\right\}\qquad \mbox{[by \eqref{eq:permute}]}\nonumber\\
=& \sup_{f\in \Lambda^* } \left\{ \mathbb{E}_{{\bf Z}'\sim (D(\mathcal{Z}))^N}\left\{\frac{1}{N}\sum_{n=1}^N\big(f({\bf z}'_{n})\right\} -\mathbb{E}_{{\bf Z}\sim (D(\mathcal{Z}))^N}\left\{\frac{1}{N}\sum_{n=1}^N \phi (f,{\bf z}_n)\big)\right\}\right\} \nonumber\\
=& \sup_{f\in \Lambda^* } \Big\{\mathbb{E}_{{\bf Z}'\sim (D(\mathcal{Z}))^N} \big\{\widehat{\mathrm{R}}'f\big\}- \mathbb{E}_{{\bf Z}\sim (D(\mathcal{Z}))^N} \big\{\mathrm{R}_\nu f \big\} \Big\} \nonumber\\
=& \sup_{f\in \Lambda^* }\Big\{ \mathrm{R}f - \mathrm{R} \phi (f,{\bf z})\Big\}.
\end{align*}
Following the above result, we also have
\begin{align*}
\Omega_\nu =&\sup_{f\in \Lambda^* } \left\{ \mathbb{E}_{{\bf Z}'\sim (D(\mathcal{Z}))^N}\left\{\frac{1}{N} \sum_{n=1}^N  f({\bf z}_n')\right\}- \mathbb{E}_{{\bf Z}\sim (D(\mathcal{Z}))^N}\left\{\frac{1}{N}\sum_{n=1}^N\int f({\bf z})  d{\rm V}\big({\bf z}|\theta({\bf z}_n)\big) \right\}\right\}\nonumber\\
=& \sup_{f\in \Lambda^* } \left\{  \int f({\bf z})  d  {\rm P}({\bf z}) - \int f({\bf z})   d \mathbb{E}_{{\bf a}\sim D(\mathcal{Z})} \big\{ {\rm V}\big({\bf z}|\theta({\bf a})\big)\big\} \right\} \nonumber\\
=&\sup_{f\in \Lambda^* } \left\{ \int f({\bf z})\;  d\,\big[ {\rm P}({\bf z}) - \mathbb{E}_{{\bf a}\sim D({\cal Z})} \big\{ {\rm V}\big({\bf z}|\theta({\bf a})\big)\big\}  \big] \right\}.
\end{align*}}
This completes the proof.\hfill $\blacksquare$


\subsection{Proof of Theorem \ref{lem:small}}

To prove Theorem \ref{lem:small}, we need the following lemma provided by \citet{li2012ten}:

\begin{lemma}\label{lem:small}
The following conditions are equivalent: for any $\alpha>0$,
\begin{enumerate}[(i)]
\item there exist $b_1,b_2>0$ such that for any $t>0$,
\begin{equation*}
  \mathbb{E} \mathrm{e}^{-t \eta_\nu^{(2)}} \leq b_1 {\rm e}^{-b_2 t^{\frac{\alpha}{1+\alpha}}};
\end{equation*}
\item there exist $c_1,c_2>0$ such that for any $\xi>0$,
\begin{equation*}
  \mathbb{P} \left\{  \eta_\nu^{(2)} \leq \xi  \right\} \leq c_1 {\rm e }^{-c_2 \xi^{-\alpha}}.
\end{equation*}
\end{enumerate}
\end{lemma}

Now, we come up with the proof of Theorem \ref{lem:small}.

{\bf Proof of Theorem \ref{lem:small}:} According to \eqref{eq:pi}, we have
\begin{align}\label{eq:small.pr1}
\mathbb{P}\{\eta_\nu<0 \} = 1- \mathbb{P}\{\eta^{(2)}_\nu\leq \eta^{(1)}_\nu \}\geq 1- \mathbb{P}\{\eta^{(2)}_\nu\leq \tau \}.
\end{align}
Then, the combination of \eqref{eq:small.pr1} and Lemma \ref{lem:small} leads to the result \eqref{eq:small.result}. This completes the proof. \hfill$\blacksquare$





\bibliography{ref-vrm}

\end{document}